\documentclass{article}
\usepackage{algorithm}
\usepackage{algpseudocode}
\usepackage{graphicx}
\usepackage{amsmath}
\usepackage{placeins}
\usepackage{multirow}
 
\usepackage[preprint]{neurips_2026}

\usepackage[utf8]{inputenc} 
\usepackage[T1]{fontenc}    
\usepackage{hyperref}       
\usepackage{url}            
\usepackage{booktabs}       
\usepackage{amsfonts}       
\usepackage{nicefrac}       
\usepackage{microtype}      
\usepackage{xcolor}         

\title{Closed-form predictive coding via hierarchical Gaussian filters}

\author{%
Aleksandrs Baskakovs \\
Center for Humanities Computing\\
Aarhus University\\
Aarhus, Denmark \\
\texttt{aleks@cas.au.dk} \\
\And
Sylvain Estebe \\
Center for Humanities Computing\\
Aarhus University\\
Aarhus, Denmark \\
\texttt{sylvainestebe@cas.au.dk} \\
\And
Kenneth Enevoldsen \\
Center for Humanities Computing\\
Aarhus University\\
Aarhus, Denmark \\
\texttt{kenneth@cas.au.dk} \\
\And
Kristoffer Nielbo \\
Center for Humanities Computing\\
Aarhus University\\
Aarhus, Denmark \\
\texttt{kln@cas.au.dk} \\
\And
Chris Mathys$^{*}$ \\
Interacting Minds Center\\
Aarhus University\\
Aarhus, Denmark \\
\texttt{chmathys@cas.au.dk} \\
\And
Nicolas Legrand$^{*}$ \\
Center for Humanities Computing\\
Aarhus University\\
Aarhus, Denmark \\
\texttt{nicolas.legrand@cas.au.dk} \\
}

\begin{document}

\maketitle

\maketitle

\begin{abstract}
  Predictive coding (PC) offers a local and biologically grounded alternative to backpropagation in the training of artificial neural networks, yet to date, it remains slower, and performance degrades sharply as network depth increases. We trace both problems to a single simplification: current PC networks fix the precision matrix to the identity, discarding precision-weighted prediction errors that the variational derivation requires to be fast, local, and Bayesian. We close this gap by expressing predictive coding networks as deep hierarchical Gaussian filters (HGFs) and restore precision-weighted message passing, yielding dynamic uncertainty estimates and Hebbian-compatible update rules at every layer. The resulting networks can simultaneously learn activations, weights, and precisions under a single free-energy objective, with no global error signal, and resolve inference without requiring iterations or automatic differentiation. On FashionMNIST, our solution approaches backpropagation in epoch-level wall-clock cost while converging in fewer epochs, and outperforms it on online, data efficiency, and concept-drift tasks. We thus establish that closed-form variational inference with online precision learning provides a tractable foundation for deep predictive coding networks, retaining biological and interpretative advantages, without requiring iterative relaxation or global error signals.
\end{abstract}

\section{Introduction}

Backpropagation (BP) underwrites modern deep learning. Yet its core assumptions, like a global error signal and strict separation of inference and learning phases, are incompatible with both current knowledge of cortical computations \citep{Whittington2019} and introduce interference \citep{Song2024}. Predictive coding (PC) \citep{Rao1999,Friston2005, Friston2009} offers a biologically grounded alternative in which each layer carries explicit prediction-error units, inference and learning both proceed by local energy minimisation, and weight updates are Hebbian \citep{Salvatori2026, vanZwol2026}. Under appropriate conditions, PC recovers the gradients computed by BP through purely local message passing \citep{Whittington2017, Millidge2022,Song2024}, making it a leading candidate for the next generation of energy-efficient, brain-inspired algorithms.

Scaling PC to modern deep architectures \citep{Pinchetti:2025, Goemaere:2025} has nonetheless been obstructed by two practical limitations. For $T$ iterations, PC is $O(T)$ times slower than BP on conventional sequential hardware, because activations must be relaxed toward equilibrium by iterative gradient descent, and performance degrades sharply with depth \citep{Pinchetti:2025}. Solutions have been proposed to circumvent this problem \citep{Qi:2025, Innocenti:2025, Goemaere:2025}, but they still require automatic differentiation or global signal propagation.

We trace both limitations to a single simplification: deep PC implementations almost universally fix the precision matrix to the identity and infer activations by iterative gradient descent. This discards the precision-weighted prediction errors that the variational derivation of PC requires \citep{Friston2009}, and with them the property that makes PC most distinctive as an algorithm, namely that in the linear case, learning under precision-weighted errors is formally equivalent to natural-gradient descent (see \citet{Millidge:2022b} Appendix B). Although learnable precisions have been widely acknowledged as a promising direction \citep{Salvatori2026}, no existing deep PC network endogenously infers them as posterior beliefs coupled to the rest of the hierarchy.

We close this gap by expressing predictive coding networks as generalised hierarchical Gaussian filters (gHGF) \citep{mathys:2011, mathys:2014, Mathys:2026, Weber2026}. The gHGF is a special case of generalised filtering, of which PC is the neurobiologically motivated instance \citep{Friston2009}, and its modular message-passing structure yields update equations that are derived analytically, are local by construction, require no automatic-differentiation scaffolding, and open the expressivity of network structures and dynamics that can be supported \citep{legrand:2024}. Inference therefore reduces to a one-shot, closed-form variational update, eliminating the iterative relaxation that obstructs both wall-clock efficiency and asynchronous deployment, and yields fast, Hebbian-compatible rules that simultaneously learn activations, weights, and precisions under a single free-energy objective.

Our contributions are:

\begin{enumerate}
    \item A reformulation of PC networks as deep hierarchical Gaussian filters that replaces iterative state inference with closed-form variational updates;
    \item State-of-the-art convergence on FashionMNIST, with wall-clock cost closer to BP and superior performance on online, continual, and concept-drift tasks;
    \item Hebbian-compatible weight updates that incorporate precision as a natural gradient weight;
    \item We characterize the role of dynamic precision learning in deep PC;
\end{enumerate}

Together, the locality, one-shot updates, and precision-driven Hebbian plasticity introduce a new framework for building deep predictive coding networks that retain biological and interpretative advantages without requiring iterative relaxation and automatic differentiation by design.

\begin{figure}[t]
  \centering
  \includegraphics[width=\linewidth]{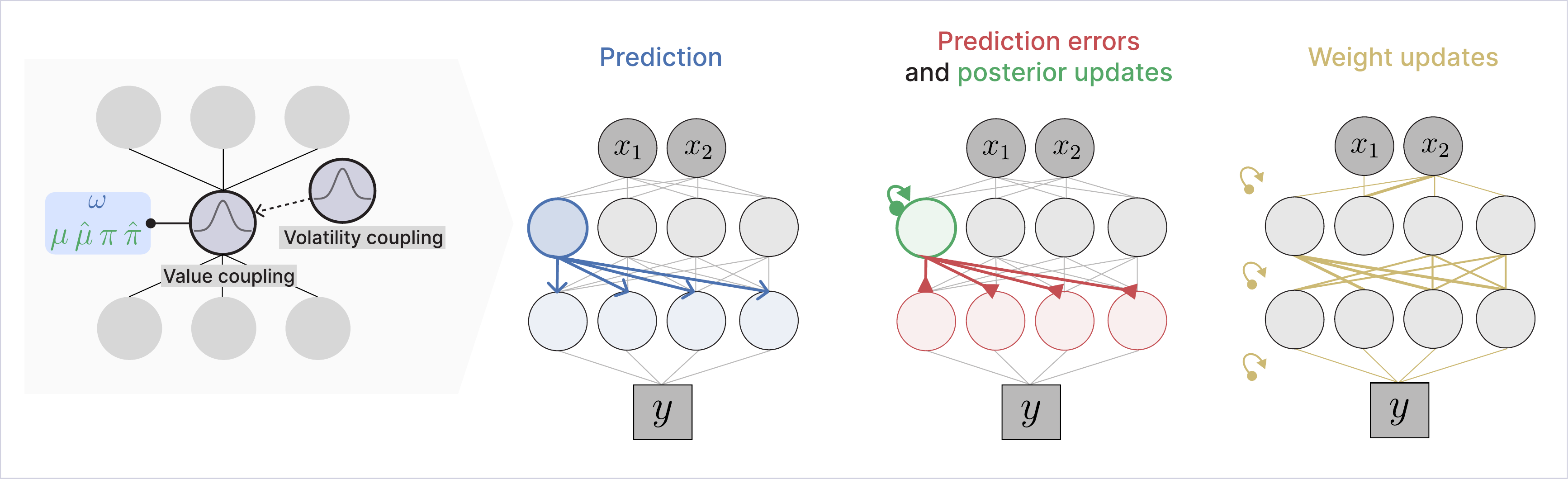}
  \caption{Predictive coding networks as deep hierarchical Gaussian filters. We express predictive coding networks as deep hierarchical Gaussian filters (HGF) where layers are connected through non-linear (here ReLU or leaky-ReLU) value coupling. An optional volatility parent can track downstream uncertainty for precision learning (left panel). While predictive coding networks rely on iterative gradient descent to minimise the local energy function, HGFs minimise the variational free energy in a single update sweep (right panel). (1) After clamping leaf nodes to predictor values, top-down predictions propagate expectations downstream, both for expected precision ($\hat{\pi}_{\ell}$) and expected mean ($\hat{\mu}_{\ell}$). (2) Bottom-up prediction errors and one-shot variational update of latent activations. (3) After the full update of latent activations, weights are adapted using a precision-weighted Hebbian update.}
  \label{fig:figure1}
\end{figure}
\section{Background and related works}

\subsection{Predictive coding networks}

Predictive coding is a family of algorithms that minimise variational free energy using different approximations and optimisation dynamics \citep{Salvatori2026, vanZwol2026}. \citet{Friston2005} recast \citeauthor{Rao1999}'s [\citeyear{Rao1999}] original predictive coding scheme as an approximate inversion of a hierarchical Gaussian generative model under the Laplace approximation, exposing PC as a special case of variational inference in which the free energy reduces to a sum of layer-wise, precision-weighted squared prediction errors, where belief updates correspond to gradient descent on this energy \citep{Bogacz2017}. Both inference and learning therefore reduce to local, Hebbian-like updates emerging from the minimisation of a single objective (i.e., "inference learning" \citep{vanZwol2026}). \citet{Whittington2017}  showed that under appropriate weight-balance conditions, inference learning approximates backpropagation in feedforward networks, a result subsequently extended to arbitrary computation graphs \citep{Millidge2022} and to networks with arbitrary topology, including recurrent, lateral, and skip connections \citep{Salvatori2022}. When activations are instead permitted to relax fully before plasticity, PC follows the distinct principle of \emph{prospective configuration} \citep{Song2024}, which empirically outperforms backpropagation on continual learning, online learning, and small-data regimes.

The framework has accordingly diversified into a family of variants targeting different applications and weaknesses: incremental schemes that reduce inference cost \citep{salvatori2024}, prevent signal decay in deep networks \citep{Goemaere:2025, Innocenti:2025}, bidirectional and hybrid models unifying generative and discriminative pathways \citep{Oliviers:2025, Tscshantz2023}, PC beyond the Gaussian assumption \citep{Pinchetti:2022}, and associative memory architectures \citep{Salvatori:2022}. Empirically, PC networks have been applied to image classification, generative modelling, and continual learning, with open-source libraries now providing standardised implementations \citep{Innocenti:2024}. Despite this breadth, these applications remain largely similar in that inference proceeds via continuous-time relaxation rather than closed-form one-step updates---therefore requiring automatic differentiation---and with precisions typically treated as fixed hyperparameters rather than as belief states updated online.

\subsection{Hierarchical Gaussian filters}

The hierarchical Gaussian filter (HGF) is a Bayesian model of perception and learning \citep{mathys:2011, mathys:2014, Mathys:2026} that casts an agent as inverting a generative model of the world in which hidden states evolve as coupled random walks. It is especially suited to inferring \textbf{volatility coupling} from experimental paradigms \citep{Sandhu2023}, whereby the value at one level sets the log-variance (i.e., step size) of the random walk one level below. That way, the subjective estimates of environmental volatility directly modulate the precision afforded to lower-level beliefs and therefore the rate at which they are updated. Approximate variational inversion under a mean-field assumption and a quadratic expansion of the variational energy yields closed-form, one-step update equations in which posterior means and precisions are revised in proportion to precision-weighted prediction errors propagated from below. A recent generalization \citep{Weber2026} and implementation \citep{legrand:2024} extend this scheme to arbitrary networks of exponential-family distributions with hierarchical, state-dependent value and volatility coupling.

This combination of analytic tractability, biologically intuitive prediction-error semantics, and individually fittable parameters has made the framework a reference in cognitive neuroscience \citep{Lawson2017, Powers2017, Mikus2025}, with open-source implementations such as PyHGF \citep{legrand:2024} distributed via TAPAS \citep{Frssle2021}. While HGF sits within the broader family of hierarchical Bayesian filtering schemes, of which predictive coding (PC) is the most prominent neurobiologically motivated instance \citep{Rao1999, Friston2005, Bogacz2017}, it differs from classic PC approaches in the form of cross-level interaction it emphasises. Classic PC is built around value coupling \citep{Whittington2017, Millidge2022, Song2024}, while the HGF's signature contribution is volatility coupling. Crucially, the two coupling types differ in how they update posterior precision: in value coupling, a parent's precision is incremented by its child's predicted precision, while volatility coupling makes precision updates an explicit function of squared prediction errors via the \emph{volatility prediction error} (VOPE) \citet{Weber2026}. While the generalised version allows for deeper versions and value couplings, no previous work has implemented generalised HGF networks for complex discrimination tasks.

\subsection{Precision and volatility learning in predictive coding}

Precision (i.e., the inverse covariance of prediction errors) occupies a theoretically central position in predictive coding. The variational free energy decomposes into a sum of layer-wise, precision-weighted squared prediction errors, and modulating precisions amounts to dynamically weighting the influence of each error signal on belief updates and learning \citep{Friston2005, Bogacz2017}. In the linear case, this weighting is formally equivalent to natural-gradient descent, with precisions playing the role of the Fisher information matrix and yielding adaptive, curvature-aware learning rates (\citep{Millidge:2022b}, Appendix B). Without it, layers cannot adaptively scale the influence of their errors, the cross-layer energy distribution becomes exponentially imbalanced \citet{Pinchetti:2025, Qi:2025}, and the network forfeits any principled handle on uncertainty.

The practical role of precision in deep PC has nonetheless remained limited. The large majority of PCNs developed for machine-learning tasks \citep{Whittington2017, Salvatori:2022, salvatori2024, Millidge2022, Pinchetti:2025, Oliviers:2025, Song2024} fix precisions to the identity throughout training, effectively assuming homoscedastic noise and reducing the free energy to a sum of unweighted squared errors. Where precision is treated as learnable, it is typically estimated as a slowly varying parameter rather than as a fast-changing belief state that is itself the target of hierarchical inference. It is acknowledged that precision-weighted Hebbian plasticity is a promising new research direction \citep{Salvatori2026}. To our knowledge, no existing deep PC network endogenously infers precisions online as posterior beliefs coupled to the rest of the hierarchy, and no systematic empirical comparison has been made between networks with fixed identity precisions and networks whose precisions are themselves the output of a hierarchical inference scheme.

\section{Method}

\subsection{Hierarchical Gaussian filters as predictive coding networks}

The computational models reported here are based on the generalised hierarchical Gaussian filter (gHGF) \citep{Weber2026}, an extension of the Hierarchical Gaussian Filter (HGF) \citep{mathys:2011, mathys:2014, Mathys:2026} to predictive coding networks that allows designing arbitrarily sized and connected networks of exponential family distributions. The gHGF tracks the posterior up to second order, updating both the mean $\mu$ and the precision $\pi$ of every belief at each step, whereas standard PC schemes optimize only the first moment and treat precisions as fixed or slowly learned hyperparameters. In a hierarchical Gaussian filter, nodes must therefore additionally transmit an estimate of prediction precision ($\hat{\pi}$) bottom-up to their parents, which has no equivalent in classical PC architectures that fix precision to a standard value like $\sigma = 1.0$. This architectural difference sets HGFs apart, as precision is not only learned and inferred but also propagates dynamically in the hierarchy as a function of the energy gradient. This also creates the possibility of having activation precision as a weight on the weight updates themselves, where standard versions only use prediction errors.

The gHGF natively supports volatility (and noise) coupling, in which a parent modulates the rate of change rather than the mean of its child via $$\hat{\pi}_a = \frac{1}{\frac{1}{\pi_a} + \exp( \mu_{\check{a}} + \omega_a)}).$$ No equivalent construction exists in standard PC, where higher levels are restricted to predicting means.

In a purely value-coupled hierarchy, the posterior precision update is independent of the prediction error: precisions accumulate as a function of bottom-up evidence and the volatility parameters $\omega_\ell$, converging to a fixed point determined by network structure rather than predictive accuracy. Modulation of precision by prediction error itself, the mechanism by which surprising observations should inflate uncertainty, requires volatility coupling \citep{Weber2026}.

Value-parent couplings alone are nonetheless sufficient to recast the gHGF as a deep network closely analogous to a PC network: stacking layers of continuous state nodes connected by nonlinear value couplings $g(\cdot)$ (e.g., ReLU or leaky ReLU as used here). The framework deviates from other PC setups in the three-step routine of predictions, prediction errors, and posterior updates (Fig. \ref{fig:figure1}).
\subsection{Generative model} 
\label{sec:gen-model}

We treat a deep network as a hierarchy of belief nodes whose joint distribution forms a generative model, which is inverted online via approximate variational inference. Following the gHGF formulation of \citet{Weber2026}, and dropping drift ($\rho_\ell = 0$) and autoconnection ($\lambda_\ell = 0$) since classification samples carry no temporal continuity in the mean, each continuous state node $\ell \in \{1,\ldots,L\}$ is generated as

\begin{equation}
x_\ell^{(k)} \sim \mathcal{N} \left(
\sum_{b\,\in\,\mathrm{vapa}(\ell)} \alpha_{b,\ell}\, g \left(x_b^{(k)}\right),\;
\exp \left(x_{\check{\ell}}^{(k)} + \omega_\ell\right) \right)
\label{eq:genmodel}
\end{equation}

where $\mathrm{vapa}(\ell)$ denotes the set of value parents of node $\ell$, $\alpha_{b,\ell}$ is the value-coupling strength from parent $b$ to node $\ell$, $\check{\ell}$ is an optional volatility parent, $g(\cdot)$ is a (leaky) ReLU nonlinearity, and $\omega_\ell$ is a tonic log-volatility parameter. When the volatility parent is absent, the variance reduces to $\exp(\omega_\ell)$, and the node behaves as a static Gaussian variable centered on its predicted mean.

Continuous input nodes at the top of the hierarchy are clamped to the data point $\mathbf{x}$. At the bottom, outcomes are represented by binary state nodes, clamped to the one-hot class label $\mathbf{y}$ during training. A binary state node generates a prediction

\begin{equation}
\hat{\mu}_{\mathrm{bin}}^{(k)} =
\sigma \left(\hat{\mu}_{pa}^{(k)}\right),
\qquad
\hat{\pi}_{\mathrm{bin}}^{(k)} =
\frac{1}{\hat{\mu}_{\mathrm{bin}}^{(k)} \left(1 - \hat{\mu}_{\mathrm{bin}}^{(k)}\right)},
\label{eq:binnode}
\end{equation}

where $\sigma$ is the logistic sigmoid and $pa$ indexes the binary node's continuous value parent. The corresponding surprise reduces to a binary cross-entropy and serves as the network's discriminative loss.

The architectural distinction of the gHGF is therefore that prediction precisions $\hat{\pi}_\ell$ are first-class belief states inferred online, rather than identity-clamped hyperparameters. Table~\ref{tab:notation} summarises the mapping. This opens new possibilities for the inference step that is central to most PC algorithms. In iterative inference learning, posterior means are obtained by gradient descent
$\dot{\mu}_\ell = -\partial F / \partial \mu_\ell$ run for $T \approx 32$--$128$
steps until equilibrium \citep{Song2024, Whittington2017, salvatori2024, Innocenti:2024}. The quadratic expansion underlying Eqs.~(\ref{eq:upd-pi})--(\ref{eq:upd-mu}) in the gHGF instead admits an analytic stationary point in a single sweep. We provide the exact equations for prediction, prediction errors, and posterior updates in \ref{sec:updates} as described in \citet{Weber2026}. We also use robust updates specific to the volatility parent as described in \citet{Mathys:2026}.

\subsection{Hebbian weight updates}
\label{sec:weights}

Once activations and precisions are updated, the value-coupling strengths
$\alpha_{b,\ell}$ (which play the role of synaptic weights; Table~\ref{tab:notation}) are revised by local Hebbian rules. We compare three rules of increasing dependence on inferred precision:

\begin{align}
\alpha^* &= \alpha + \eta\, \delta_\ell\, g \left(\mu_{\ell+1}\right),
\label{eq:w-std}\\
\alpha^* &= \alpha + \eta\, \delta_\ell\, g \left(\mu_{\ell+1}\right)
\cdot \pi_{\ell}
\label{eq:w-prec}\\
\alpha^* &= \alpha + \eta\, \delta_\ell\, g \left(\mu_{\ell+1}\right) \cdot
\frac{\hat{\pi}_\ell}{\hat{\pi}_\ell + \hat{\pi}_{\ell+1}}.
\label{eq:w-ratio}
\end{align}

Eq.~(\ref{eq:w-std}) is the standard rule used in predictive coding networks with fixed identity precision \citep{Whittington2017, Song2024}. Eq.~(\ref{eq:w-prec}) is the same Hebbian rule with a multiplicative precision factor that furnishes a learning rate per connection. Eq.~(\ref{eq:w-ratio}) additionally normalises by the parent's posterior precision at the level above, preventing runaway gain when an upstream node becomes highly certain. We use Eq.~(\ref{eq:w-prec}) in the main benchmark and Eq.~(\ref{eq:w-ratio}) to test the role of precision learning to mitigate vanishing gradients.

\paragraph{Network structures.}

We compare two networks structures, all sharing the prediction--update--PE routine of \S\ref{sec:updates}:

\begin{enumerate}
    \item \textbf{Temporal with precision propagation.} Samples are
    processed sequentially; $\mu_\ell$ are given by the predictors, but $\pi_\ell$ is inherited via successive bottom-up posterior updates. Inherited precision competes with tonic volatility $\omega_\ell$ at the prediction step (Eq.~\ref{eq:pred-pi}), but updates remain prediction-error-independent.
    \item \textbf{Temporal with precision propagation and implicit volatility parents.} Processing like in (1), but now each state node has a volatility parent whose posterior is updated by squared prediction errors, introducing the channel through which prediction errors modulate precision.
\end{enumerate}

\subsection{Implementation}
\label{sec:software}

All models compared in \S\ref{sec:experiments} were implemented in JAX \citep{Bradbury:2018} to ensure a fair comparison across frameworks. Predictive coding networks were implemented using JPC
\citep{Innocenti:2024}. Deep neural networks trained by backpropagation were implemented in Equinox v0.13.7 \citep{Kidger:2021}. Deep hierarchical Gaussian filters were implemented in PyHGF \citep{legrand:2024} v0.2.11, using the vectorised JAX backend that scales to deep networks.

\section{Experiments}
\label{sec:experiments}

We evaluate deep hierarchical Gaussian filters (HGF) against multilayer perceptrons trained with backpropagation (MLP) and vanilla predictive coding networks on FashionMNIST \citep{Xiao:2027} with a suite of supervised and online learning benchmarks used by \citep{Song2024}. We restrict comparison to baselines that preserve PC's core commitments: local computation, a single free-energy objective, and no global error signal. We accordingly benchmark against the canonical PCN \citet{Whittington2017} and BP, which provide the principled baselines for the comparisons made here.

\begin{figure}[t]
  \centering
  \includegraphics[width=\linewidth]{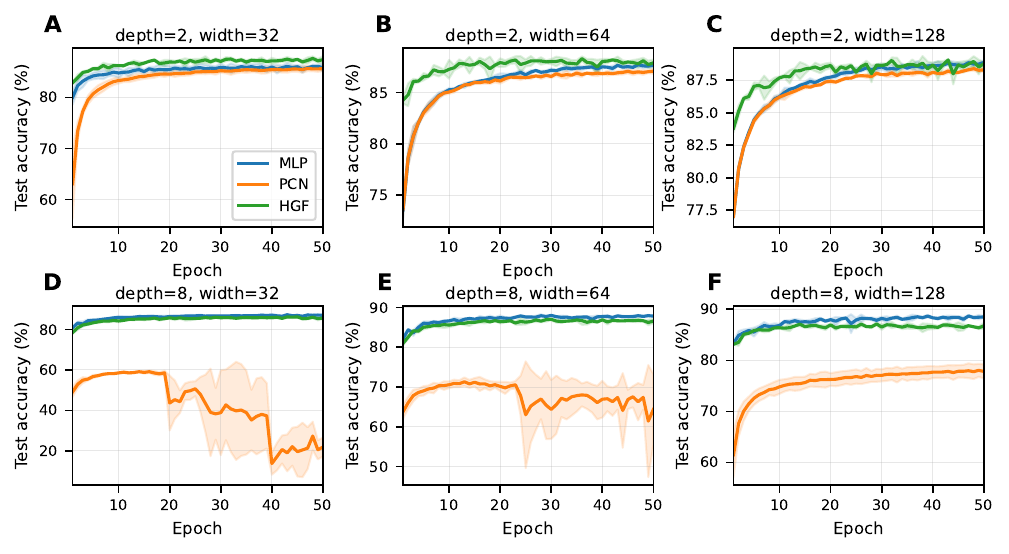}
  \caption{
  Test accuracy learning curves on FashionMNIST. Rows correspond to hidden depth ($d \in \{2, 8\}$), columns to hidden width ($w \in \{32, 64, 128\}$). Shaded regions denote $\pm 1$ standard deviation across 3 seeds.  Learning rates selected by the oracle criterion per method and configuration.
  }
  \label{fig:2}
\end{figure}

\paragraph{Shared architecture.}
All models use a fully-connected architecture with leaky ReLU activations (negative slope $0.01$) and bias terms at every layer. We sweep hidden depth $d \in \{2, 8\}$ and width $w \in \{32, 64, 128\}$ to assess how each method scales with model capacity. Weights are initialised with He initialisation, and performance is reported as the oracle-best learning rate (the LR maximising mean test accuracy across seeds), averaged over 3 random seeds.

\paragraph{Network configurations.}
The MLP is optimised with Adam over learning rates $\{10^{-2}, 10^{-3}, 10^{-4}\}$ with batch size 64. The PC networks perform $T = 20$ inference steps per sample using SGD on activities (learning rate $0.1$), followed by an Adam weight update with the same learning rate sweep and batch size as the MLP. In the main experiments on speed and performance, we disabled precision learning through volatility coupling in the HGF. Tonic volatility is set to $\omega = -10$ and learning rates are swept over $\{10^{-4}, 5{\times}10^{-4}, 10^{-3}, 2{\times}10^{-3}\}$.

\begin{figure}[t]
  \centering
  \includegraphics[width=\linewidth]{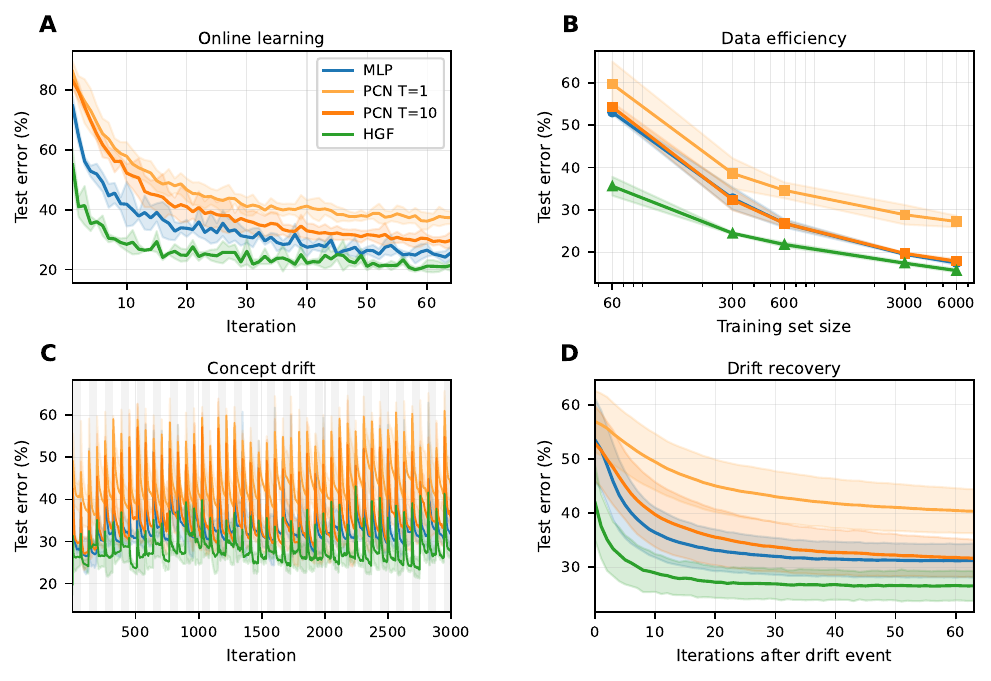}
  \caption{Adaptation and efficiency benchmarks. \textbf{(a)}~Online learning:
           test error vs.\ iteration (batch size 1, 200 samples per iteration).
           \textbf{(b)}~Data efficiency: test error vs.\ training set size
           (64 epochs per run).
           \textbf{(c)}~Concept drift: test error over 3{,}000 iterations;
           label mapping for classes 5--9 permuted every 64 iterations
           (alternating grey bands).
           \textbf{(d)}~Drift recovery: mean test error aligned to drift events,
           averaged over ${\sim}47$ events and 3 seeds.
           All panels use oracle-best learning rate per method; shaded regions
           denote $\pm 1$ std across seeds.}
  \label{fig:3}
\end{figure}

\subsection{Direct comparison}
\label{sec:direct_comparison}

All methods are trained on the full FashionMNIST training set (60{,}000 samples) for 50 epochs and evaluated on the 10{,}000-sample test set. Figure~\ref{fig:2} shows test accuracy learning curves across all depth-width configurations, and Table~\ref{tab:classification} reports oracle-best accuracy for each setting.

\subsection{Online learning}
\label{sec:online_learning}

We evaluate all methods in a strictly online regime (batch size 1) over 64 iterations, each consisting of 200 randomly drawn FashionMNIST samples. Test error is recorded after every iteration, and learning rates are chosen by the oracle criterion. All models use a fixed architecture of two hidden layers of width 32 for comparability.

Panel~(a) of Figure~\ref{fig:3} shows test error over iterations. The HGF converges faster and to a lower steady-state error than both baselines across all 64 iterations.

\begin{figure}[t]
  \centering
  \includegraphics[width=\linewidth]{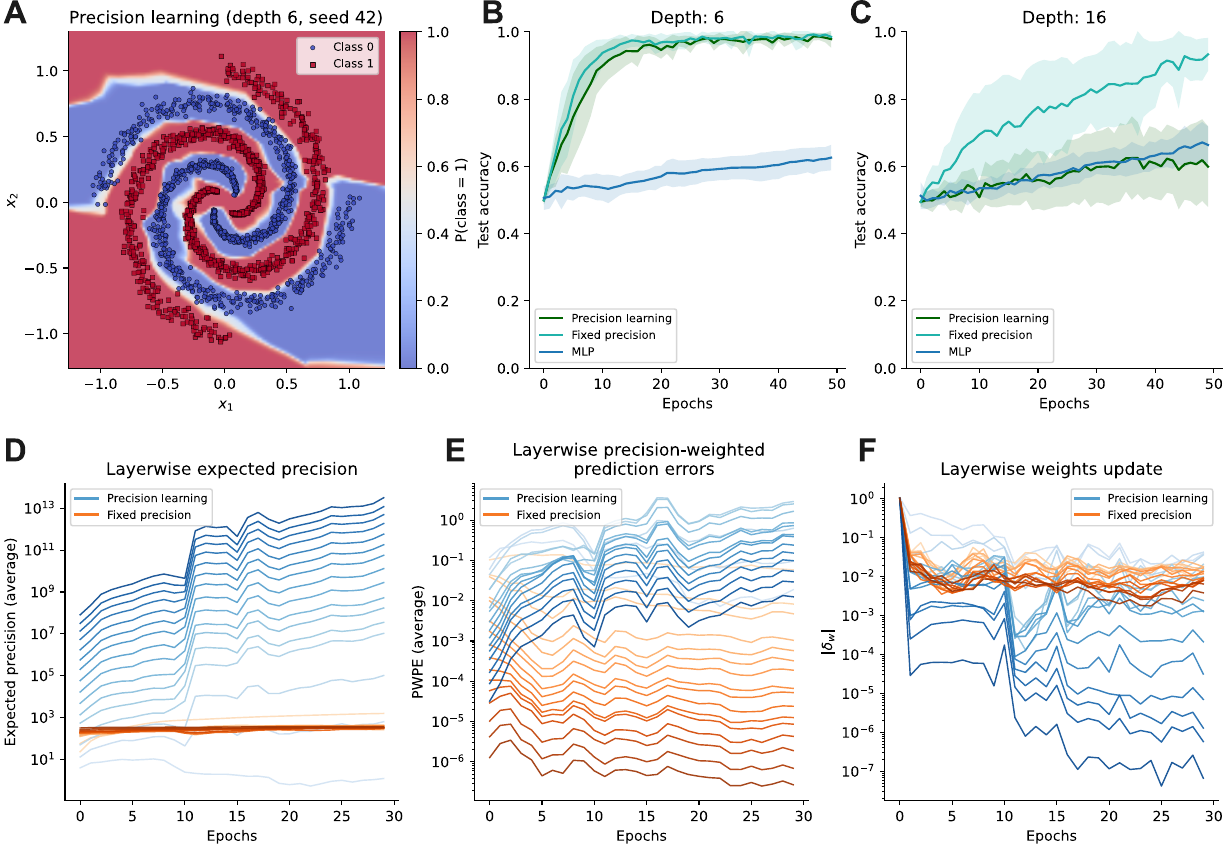}
  \caption{Activation precision as a gating mechanism for learning and inference in deep networks. \textbf{(a)} We tested two types of networks (with and without precision learning, here reporting the post-training prediction space for a network with precision learning) on a 4-arm spiral classification task to inspect both learning and inference dynamics. Both structures were able to solve the task efficiently in less than 20 epochs for depth-6 \textbf{(b)} and showed learning dynamics in depth-16 with an advantage for non-volatile structures in both cases. Inspecting the averaged states trajectories, we observed that networks equipped with volatility coupling (Blues) increased their expected precision sharply by multiple orders of magnitude in a hierarchical manner, such as the deepest layer ended up being more precise \textbf{(d)}. Networks without such learning dynamics maintained stable precision expectation with no layer differences (Oranges). This difference converted into more stable precision-weighted prediction errors \textbf{(e)}, but did not favour the stability of weight updates \textbf{(f)}.}
  \label{fig:4}
\end{figure}


\subsection{Data efficiency}
\label{sec:small_data}

We assess data efficiency by training all models on subsets of FashionMNIST of sizes $n \in \{60, 300, 600, 3000, 6000\}$, evaluating on the full held-out test set. Each model is trained for 64 epochs per run; oracle-best accuracy across the learning rate sweep is reported.

Panel~(b) of Figure~\ref{fig:3} shows test error as a function of training set size. HGF achieves lower error than both baselines at small dataset sizes, with the gap narrowing as $n$ increases.


\subsection{Concept drift}
\label{sec:concept_drift}

We evaluate adaptation to non-stationary environments using a concept drift protocol. Models are first pre-trained online on 6{,}000 samples for 64 epochs. During the drift phase, 3{,}000 iterations are run; every 64 iterations the label
mapping for classes 5--9 is randomly permuted, simulating abrupt concept drift, while classes 0--4 retain fixed labels. Each iteration uses 120 training and 1{,}000 test samples. Precision estimates in the HGF are \emph{not} reset between
drift events, so the model must adapt using its accumulated belief state.

Panels~(c) and~(d) of Figure~\ref{fig:3} show the full test-error time series and the mean post-drift recovery curve (averaged over ${\sim}47$ drift events and 3 seeds), respectively. The HGF maintains a lower steady-state error throughout the drift sequence and recovers more rapidly following each drift event.

\subsection{Precision learning}

We evaluate the impact of precision learning on training stability and accuracy in deep predictive-coding networks using a synthetic 4-arm spiral binary classification task with an 80/20 train/test split. Three architectures are compared at depth 6 and depth 16 (width 12 throughout): (i) a PyHGF DeepNetwork with precision learning, in which every hidden layer carries an implicit volatility parent; (ii) a fixed-precision PyHGF baseline; and (iii) a standard multilayer perceptron with leaky-ReLU activations. We use He weight initialisation and leaky-RELU activation function across all three architectures. The MLP is trained by sigmoid cross-entropy with Adam ($ lr = 5 \times 10^{-3} $). All networks are trained for 50 epochs in full-batch mode across 10 random seeds. The reported metric is test classification accuracy, summarised as mean $\pm 1$ standard deviation across seeds.

To characterise the mechanisms underlying the architecture differences, we examined the layerwise trajectories from the depth-16 precision-learning and fixed-precision PyHGF networks, providing a direct readout of effective gradient flow across depth. For each volatile layer i we compute the mean expected precision $\hat{\pi}_i$, capturing the accumulation/dissipation of precision during training, and the mean precision-weighted prediction error $\mathrm{PWPE}_i = |\mu_i - \hat{\mu}_i| \cdot \hat{\pi}_i$, the magnitude of the upward error signal that drives belief and weight updates. We additionally record the mean absolute weight change $|\Delta W_i|$ to measure learning-related updates.

\section{Conclusion and limitations}
\label{sec:limitations}

We have focused on establishing the algorithm's performance along the most central axes to predictive coding: locality of updates, absence of a global error signal, and the Bayesian machinery that enables dynamic precision inference. Our comparison is restricted to baselines preserving these commitments. Recent depth-scaling proposals such as ePC \citep{Goemaere:2025} and $\mu$PC \citep{Innocenti:2025} address related concerns through complementary mechanisms whose design choices lie outside this regime. A controlled comparison would clarify the trade-off between fidelity to the variational derivation and the engineering choices that enable scaling, but this lies beyond the present scope.

Following \citet{Song2024}, we evaluated on FashionMNIST \citep{Xiao:2027}, a standard benchmark with sufficient complexity to differentiate methods, but one that does not establish the scaling potential of the approach to larger natural-image datasets. A true test of depth would require convolutional architectures applied to tasks such as CIFAR or TinyImageNet; extending the current HGF framework to support such architectures is a natural next step, but lies outside the present scope.

The most substantive open question concerns the role of precision learning in enabling and stabilising inference and learning in PC networks. On a synthetic small-scale task, volatility parents successfully stabilised precision-weighted prediction errors and produced the hierarchical precision structure expected from the variational derivation (Fig.~\ref{fig:4}D-E). This comes at the cost of a growing precision gap between layers that introduces its own imbalance in inference and learning. A Hebbian rule tailored to this regime (a complementary Kalman gain) only partially compensates: at depth 16, the fixed-precision baseline still outperforms the volatility-coupled variant despite the latter's better-conditioned error signals. Deriving weight-update rules consistent with volatility-coupled precision dynamics is the principal next step opened by this work.

Finally, our implementation samples sequentially rather than in batches, which alone accounts for most of the wall-clock gap to BP at deep configurations (Fig.~\ref{fig:5}). A vectorised batched form admitted by the closed-form updates is left to future work.

\FloatBarrier


\bibliographystyle{plainnat}
\bibliography{references}

\newpage
\appendix
\section{Appendix}

\subsection{Closed-form variational updates}
\label{sec:updates}

\paragraph{Prediction step (top-down)}

Given posterior beliefs from the previous trial, each node generates a new mean and precision:

\begin{align}
\hat{\mu}_\ell^{(k)} &=
\sum_{b\,\in\,\mathrm{vapa}(\ell)} \alpha_{b,\ell}\, g \left(\mu_b^{(k)}\right),
\label{eq:pred-mean}\\[2pt]
\hat{\pi}_\ell^{(k)} &=
\frac{1}{\dfrac{1}{\pi_\ell^{(k-1)}}
+ \exp \left(\mu_{\check{\ell}}^{(k)} + \omega_\ell\right)}.
\label{eq:pred-pi}
\end{align}

For the batch-independent regime, $\pi_\ell^{(k-1)}$ is fixed between samples. For temporal regimes, $\pi_\ell^{(k-1)}$ is carried over, allowing posterior precision to accumulate across samples.

\paragraph{Posterior updates}

For value parent $b$ of child $a$, we adapt Eqs.~50--51 of \citet{Weber2026} by evaluating the derivatives at the prediction $\hat{\mu}_b^{(k)}$ rather than the previous-trial posterior $\mu_b^{(k-1)}$. This reflects the fact that, with $\rho_\ell = \lambda_\ell = 0$, samples are not autoregressive in the mean, and the prediction is the appropriate expansion point:

\begin{align}
\pi_b^{(k)} &= \hat{\pi}_b^{(k)}
+ \hat{\pi}_a^{(k)}  \left(
\alpha_{b,a}^2\, g' \left(\hat{\mu}_b^{(k)}\right)^{ 2}
- \alpha_{b,a}g'' \left(\hat{\mu}_b^{(k)}\right) \delta_a^{(k)}
\right),
\label{eq:upd-pi}\\[2pt]
\mu_b^{(k)} &= \hat{\mu}_b^{(k)}
+ \frac{\hat{\pi}_a^{(k)}\, \alpha_{b,a}\, g' \left(\hat{\mu}_b^{(k)}\right)}{\pi_b^{(k)}}\,
\delta_a^{(k)}.
\label{eq:upd-mu}
\end{align}

For piecewise-linear $g$ (ReLU, leaky ReLU), $g''(\cdot) = 0$ everywhere, so the precision update~(\ref{eq:upd-pi}) collapses to a non-negative term and the mean update~(\ref{eq:upd-mu}) becomes a precision-weighted Kalman gain on the prediction error.

\paragraph{Prediction error step}

Each node sends to its parents from the layer above:

\begin{equation}
\delta_a^{(k)} = \mu_a^{(k)} - \hat{\mu}_a^{(k)},
\qquad \text{together with } \hat{\pi}_a^{(k)}.
\label{eq:vape}
\end{equation}

\subsection{Notation and mapping between gHGF and standard PC quantities.}
\label{sec:mapping}

\begin{table}[H]
\caption{Notation and mapping between gHGF and standard PC quantities.}
\label{tab:notation}
\centering
\small
\begin{tabular}{lll}
\toprule
\textbf{Symbol} & \textbf{gHGF role} & \textbf{PC equivalent} \\
\midrule
$\mu_\ell, \hat{\mu}_\ell$       & posterior / predicted mean of node $\ell$    & $\mu_\ell$, $f(W \mu_{\ell+1})$ \\
$\pi_\ell, \hat{\pi}_\ell$       & posterior / predicted precision of node $\ell$ & $\Sigma_\ell^{-1}$ (typically fixed) \\
$\delta_\ell$                    & value prediction error (VAPE)                & $\epsilon_\ell$ \\
$\alpha_{b,\ell}$                & value-coupling strength parent$\to$child     & $W_{\ell+1,\ell}$ \\
$\omega_\ell$                    & tonic log-volatility                         & typically fixed to identity \\
$g(\cdot)$                       & nonlinear value coupling                     & activation function $f$ \\
\bottomrule
\end{tabular}
\end{table}

\subsection{Algorithms}

\begin{algorithm}
\caption{Deep Hierarchical Gaussian Filter --- single training step}
\textbf{Input:} input $x$, target $y$, network $\mathcal{N}$
\begin{algorithmic}[1]
\State $\mu_L \gets x$, \quad $\mu_0 \gets y$

\vspace{0.3em}
\State \textbf{Prediction}
\For{$\ell = L-1$ \textbf{down to} $0$}
    \State $\hat{\mu}_\ell \gets \alpha_\ell\, g(\mu_{\ell+1})$
    \State $\hat{\pi}_\ell \gets \dfrac{1}{\frac{1}{\pi_\ell} + \exp(\omega_\ell)}$
\EndFor

\vspace{0.3em}
\State \textbf{Prediction error}
\For{$\ell = 0$ \textbf{to} $L-1$}
    \State $\delta_\ell \gets \mu_\ell - \hat{\mu}_\ell$
\EndFor

\vspace{0.3em}
\State \textbf{Posterior update}
\For{$\ell = 1$ \textbf{to} $L-1$}
    \State $\pi_\ell \gets \hat{\pi}_\ell + \hat{\pi}_{\ell-1}\, \alpha_{\ell-1}^2\, g'(\hat{\mu}_\ell)^2$
    \State $\mu_\ell \gets \hat{\mu}_\ell + \dfrac{\hat{\pi}_{\ell-1}\, \alpha_{\ell-1}\, g'(\hat{\mu}_\ell)}{\pi_\ell}\, \delta_{\ell-1}$
\EndFor

\vspace{0.3em}
\State \textbf{Weight update}
\For{$\ell = 0$ \textbf{to} $L-1$}
    \State $\alpha_\ell \gets \alpha_\ell + \eta\, \dfrac{\pi_\ell}{\pi_\ell + \pi_{\ell+1}}\, \delta_\ell\, g(\mu_{\ell+1})$
\EndFor
\end{algorithmic}
\end{algorithm}

\subsection{Computational cost}
\label{sec:computational_cost}

\begin{figure}[t]
  \centering
  \includegraphics[width=\linewidth]{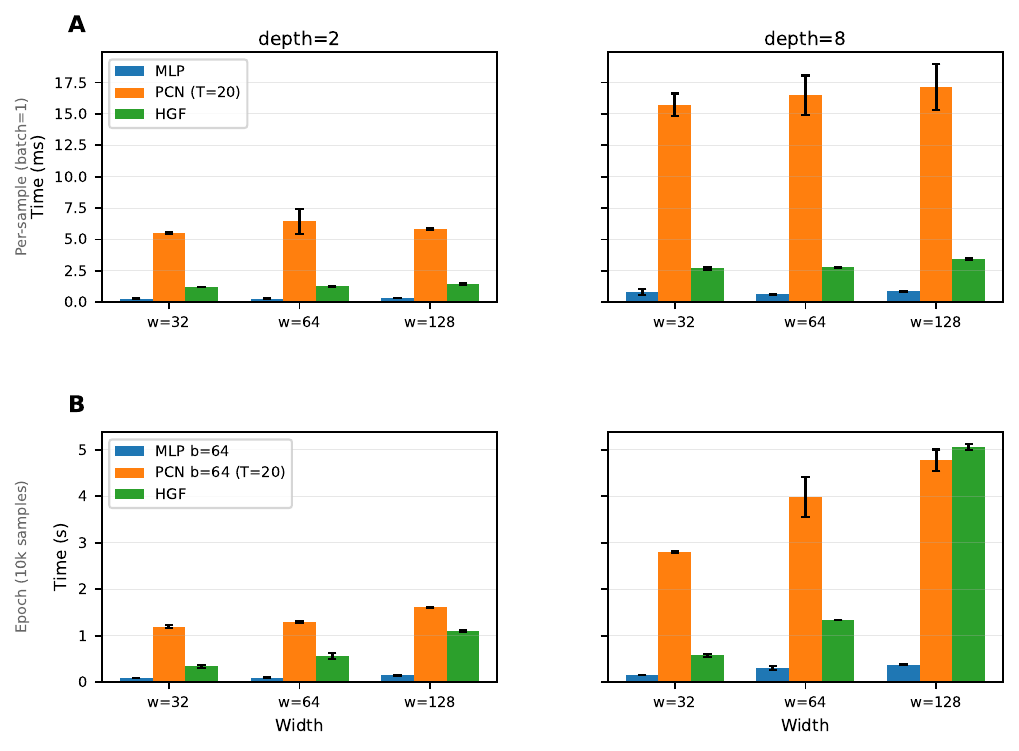}
  \caption{Computational cost comparison across depth and width configurations.
           \textbf{Top:} per-sample wall-clock time (batch size 1, 100 timed trials
           after 20 warm-up trials).
           \textbf{Bottom:} epoch wall-clock time over 10{,}000 FashionMNIST samples
           (MLP and PCN at batch size 64; HGF processes samples sequentially).
           Error bars denote $\pm 1$ standard deviation.}
  \label{fig:5}
\end{figure}

We report wall-clock times for single-sample updates and full epochs over 10{,}000 samples; none of the implementations were optimised for throughput (Figure~\ref{fig:5}).  At per-sample, MLP is fastest ($0.28$--$0.85$\,ms), HGF is $4$--$5\times$ slower ($1.21$--$3.43$\,ms), and PCN is the most expensive ($5.51$--$17.13$\,ms).  At the epoch level, HGF processes samples sequentially and currently lacks a batched inference path; even so, it remains competitive with PCN at batch size 64 ($0.34$--$5.05$\,s vs.\ $1.19$--$4.77$\,s) across most configurations, and at shallow depths approaches
MLP in wall-clock time ($0.34$--$1.10$\,s vs.\ $0.08$--$0.14$\,s at depth~2). HGF's overhead is implementation-bound rather than algorithmic: extending the update to support batched inputs is a natural next step and would be expected to substantially reduce the gap to MLP.

\subsection{Oracle-best test accuracy}

\begin{table}[H]
  \centering
  \caption{Oracle-best test accuracy (\%) on FashionMNIST. Mean across 3 seeds;
           best learning rate per (method, depth, width).
           \textbf{Bold} marks the best method per row.}
  \label{tab:classification}
  \setlength{\tabcolsep}{9pt}
  \begin{tabular}{ccrrr}
    \toprule
    Depth & Width & MLP   & PCN   & HGF   \\
    \midrule
    \multirow{3}{*}{2}
      & 32  & 86.31 & 85.70 & \textbf{88.00} \\
      & 64  & 87.84 & 87.14 & \textbf{88.51} \\
      & 128 & 88.98 & 88.43 & \textbf{89.22} \\
    \midrule
    \multirow{3}{*}{8}
      & 32  & \textbf{87.43} & 59.34 & 86.42 \\
      & 64  & \textbf{88.23} & 71.56 & 87.31 \\
      & 128 & \textbf{88.83} & 78.13 & 87.31 \\
    \bottomrule
  \end{tabular}
\end{table}


\end{document}